\documentclass[runningheads]{llncs}

\usepackage{eccv}

\usepackage{eccvabbrv}

\usepackage{graphicx}
\usepackage{booktabs}

\usepackage[accsupp]{
    axessibility
} %

\usepackage[pagebackref, breaklinks, colorlinks, citecolor=eccvblue]{hyperref}

\usepackage{orcidlink}

\usepackage{amsmath,amsfonts,bm}

\def\eqref#1{equation~\ref{#1}}

\def\1{\bm{1}}

\DeclareMathAlphabet{\mathsfit}{\encodingdefault}{\sfdefault}{m}{sl}
\SetMathAlphabet{\mathsfit}{bold}{\encodingdefault}{\sfdefault}{bx}{n}

\definecolor{cvprblue}{rgb}{0.21,0.49,0.74}

\usepackage{graphicx}
\usepackage{xspace}
\usepackage{booktabs}
\usepackage{wrapfig}
\usepackage{array}
\usepackage{tikz}
\usepackage{pgfplots}
\usepackage{colortbl}
\usepackage[accsupp]{axessibility}  %
\usepackage{orcidlink}
\usepackage{tabularx}
\usepackage{multirow}
\usepackage{tcolorbox}
\usepackage{tikz}
\usepackage{pifont}
\usepackage{microtype}
\usepackage{fontawesome}
\usepackage[toc,page]{appendix}
\usepackage{bm}
\usepackage{gradient-text}
\usepackage{textcomp}
\usepackage{gensymb}
\usepackage{lipsum}
\usepackage{enumitem}
\usepackage{makecell}

\usepackage{caption}
\usepackage[capitalize]{cleveref}
\crefname{section}{Sec.}{Secs.}
\Crefname{section}{Section}{Sections}
\Crefname{table}{Table}{Tables}
\crefname{table}{Tab.}{Tabs.}

\usepackage{caption}
\newcommand{\qheading}[1]{\noindent\mbox{\textbf{#1}}}
\newcommand{\pheading}[1]{\medskip\noindent\textbf{#1}}

\definecolor{bestgreen}{RGB}{153,200,76}
\definecolor{worstred}{RGB}{192,0,0}

\newcommand{\best}{\cellcolor{red!25}} 
\newcommand{\second}{\cellcolor{orange!25}}

\definecolor{cbad}{HTML}{FFD0D0}
\definecolor{cmedium}{HTML}{FFF0D0}
\definecolor{cgood}{HTML}{90C060}

\usepackage{pgfplots}
\pgfplotsset{compat=1.18}

\newlength{\savewidth}

\definecolor{DeltaColor}{rgb}{0.039,0.73,0.71}
\definecolor{SigmaColor}{rgb}{0.98,0.45,0.0}
\definecolor{AlphaColor}{rgb}{0,0,0.8}
\definecolor{BetaColor}{rgb}{0.8,0,0.8}
\definecolor{GammaColor}{rgb}{0.514,0.34,0.224}
\definecolor{EpsilonColor}{rgb}{0.353,0.725,0.906}
\definecolor{PurpleColor}{HTML}{9839ff}
\definecolor{RedColor}{rgb}{0.949,0.275, 0.224}
\definecolor{citecolor}{HTML}{0071bc}

\definecolor{deepred}{HTML}{940000}

\newcommand{\page}{\href{https://xiaobenli00.github.io/ETCH-X}{\textcolor{magenta}{\xspace\texttt{xiaobenli00.github.io/ETCH-X}}}\xspace}

\newcommand{\modelname}{\mbox{{ETCH-X}}\xspace}

\newcommand{\longtitle}{Robustify Expressive Body Fitting to Clothed Humans with Composable Datasets}
\newcommand{\ourtitle}{\modelname: \longtitle}

\newcommand{\xmark}{\textcolor{RedColor}{\ding{55}}\xspace}
\newcommand{\cmark}{\textcolor{GreenColor}{\ding{51}}\xspace}

\newcommand{\gt}{{ground-truth}\xspace}

\newcommand{\arteq}{\mbox{ArtEq}\xspace}
\newcommand{\ipnet}{\mbox{IPNet}\xspace}

\newcommand{\ddress}{\mbox{4D-Dress}\xspace}
\newcommand{\cape}{\mbox{CAPE}\xspace}
\newcommand{\nicp}{\mbox{NICP}\xspace}
\newcommand{\ptf}{\mbox{PTF}\xspace}
\newcommand{\etch}{\mbox{ETCH}\xspace}

\newcommand{\clothd}{\mbox{CLOTH3D}\xspace}
\newcommand{\amass}{\mbox{AMASS}\xspace}
\newcommand{\interhand}{\mbox{InterHand2.6M}\xspace}
\newcommand{\bedlam}{\mbox{BEDLAM2.0}\xspace}

\newcommand{\smplx}{\mbox{SMPL-X}\xspace}
\newcommand{\smpl}{\mbox{SMPL}\xspace}

\newcommand{\sota}{state-of-the-art\xspace}

\newcommand{\real}{\mathbb{R}}
\newcommand{\vect}[1]{\mathbf{#1}}

\newcommand{\shapecoeff}{\boldsymbol{\beta}}
\newcommand{\expressioncoeff}{\boldsymbol{\psi}}
\newcommand{\shapedim}{{\left| \shapecoeff \right|}}

\newcommand{\posecoeff}{\boldsymbol{\theta}}

\newcommand{\template}{\mathbf{\bar{T}}}
\newcommand{\restpose}{\mathbf{T}}

\newcommand{\blendweights}{\mathcal{W}}

\definecolor{PurpleColor}{HTML}{8B008B}
\definecolor{OrangeColor}{rgb}{0.914,0.541,0.0.141}
\definecolor{GreenColor}{rgb}{0.137,0.573,0.565}

\begin{document}
    \title{\ourtitle}

\author{
  Xiaoben Li\inst{1,2,3} \and
  Jingyi Wu\inst{2,4} \and
  Zeyu Cai\inst{2,5} \and
  Siyuan Yu\inst{2} \and \\
  Boqian Li\inst{2} \and
  Yuliang Xiu\inst{2}\thanks{Corresponding author.}
}
\authorrunning{X. Li, J. Wu, Z. Cai, S. Yu, B. Li and Y. Xiu}
\titlerunning{\modelname}
\institute{
  Zhejiang University\and
  Westlake University \and
  Shanghai Innovation Institute \and
  Fudan University \and
  Nanjing University\\
  \email{
    \{lixiaoben, yusiyuan, xiuyuliang\}@westlake.edu.cn,
    jingyiwu23@fudan.edu.cn,
    caizeyu010612@gmail.com,
    boqianlihuster@gmail.com
  }
  \page
}

    \maketitle

    \begin{abstract}
    Human body fitting, which aligns parametric body models, such as \smpl, to raw 3D point clouds of clothed humans, serves as a crucial first step for downstream tasks like animation and texturing.
    An effective fitting method should be both \textbf{locally expressive} -- capturing fine details such as hands and facial features -- and \textbf{globally robust} to handle real-world challenges, including clothing dynamics, pose variations, and noisy or partial inputs.
    Existing approaches typically excel in only one aspect, lacking an all-in-one solution.
    We upgrade ETCH to \modelname, which leverages a tightness-aware fitting paradigm to filter out clothing dynamics (``undress''), extends expressiveness with \smplx, and replaces explicit sparse markers (which are highly sensitive to partial data) with implicit dense correspondences (``dense fit'') for more robust and fine-grained body fitting.
    Our disentangled ``undress'' and ``dense fit'' modular stages enable separate and scalable training on composable data sources, including diverse simulated garments (\clothd), large-scale full-body motions (\amass), and fine-grained hand gestures (\interhand), improving outfit generalization and pose robustness of both bodies and hands.
    Our approach achieves robust and expressive fitting across diverse clothing, poses, and levels of input completeness, delivering a substantial performance improvement over ETCH on both 1) seen data, such as \ddress (MPJPE-All, \textcolor{GreenColor}{$33.0\%\downarrow$}) and CAPE (V2V-Hands, \textcolor{GreenColor}{$35.8\%\downarrow$}), and 2) unseen data, such as \bedlam (MPJPE-All, \textcolor{GreenColor}{$80.8\%\downarrow$}; V2V-All, \textcolor{GreenColor}{$80.5\%\downarrow$}).
    Code and models will be released at \page.
    \keywords{Clothed humans \and Partial scans \and 3D Body fitting \and Hand pose estimation \and Dense correspondences}
\end{abstract}

\begin{figure}
    \centering
    \includegraphics[width=\linewidth]{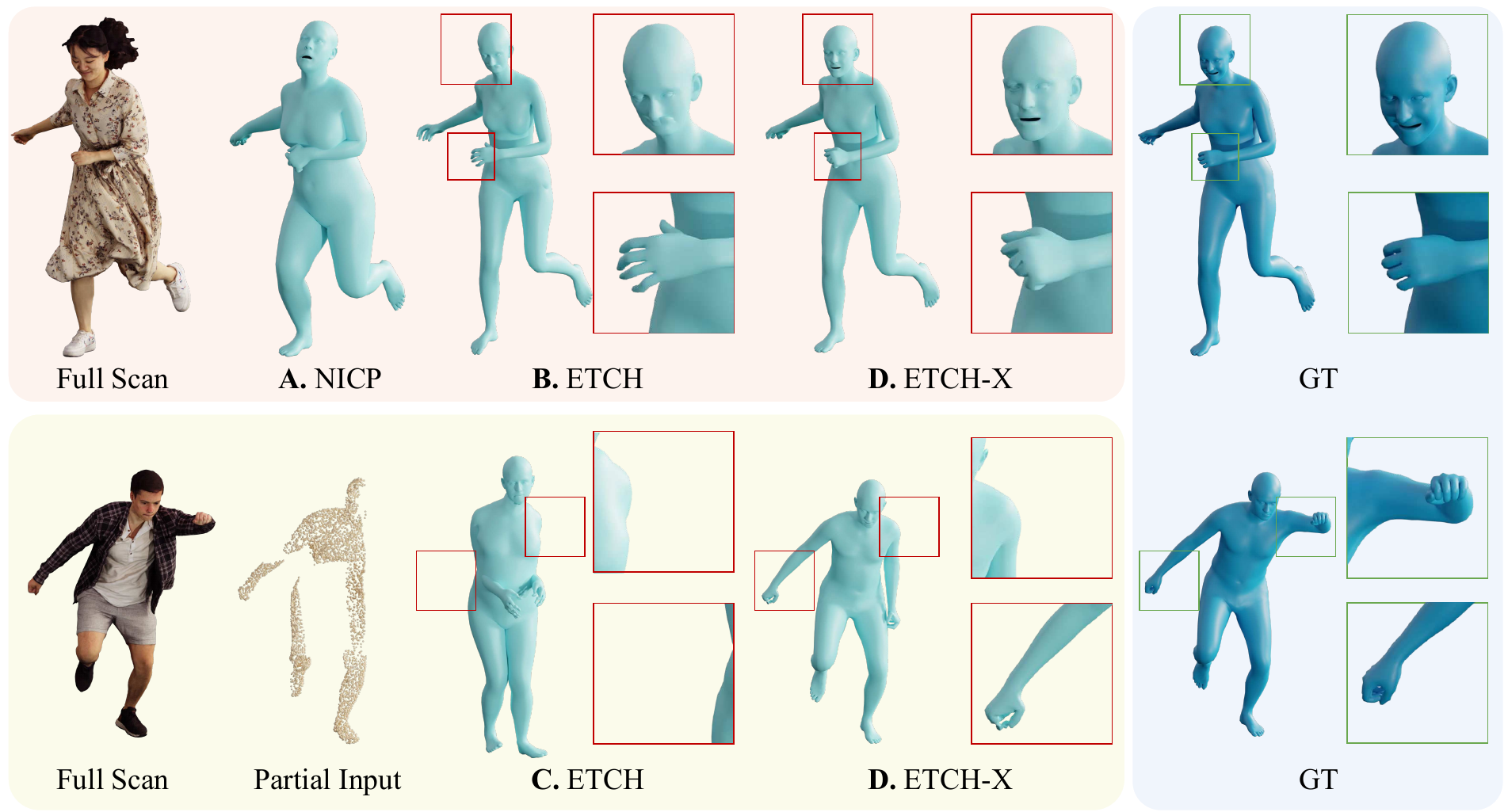}
    \caption{\textbf{Strengths of \modelname.} While NICP~\cite{marin24nicp},
    which uses implicit dense correspondence but lacks tightness-aware
    undressing, consistently produces overweight bodies from clothed scans (\textbf{A}),
    ETCH~\cite{li2025etch}, with tightness-aware undressing but sparse markers, fails
    to capture detailed body parts such as hands and face (\textbf{B}), and struggles
    with partial inputs due to missing markers (\textbf{C}). In contrast, our
    \modelname combines the strengths of both approaches, achieving robust and expressive
    fitting across diverse clothing, poses, and levels of input completeness (\textbf{D}).}
    \label{fig:teaser}
    \vspace{-2.0em}
\end{figure}

    \section{Introduction}
\label{sec:intro}

Humans are commonly captured as point clouds using 3D scanners or depth sensors.
Such point clouds are often noisy, incomplete, and lack topological structure,
preventing direct use in downstream tasks, such as shape analysis~\cite{allen2003space, cai2025up2you},
animation~\cite{luo2024smplolympics,shao2025degas}, garment refitting~\cite{de2020garment, li2025garmentdreamer},
and human interactions~\cite{huang2022rich}. A crucial step to enable these applications
is to align a parametric human body model (\eg, \smplx~\cite{SMPLX:2019}, GHUM~\cite{xu2020ghum})
to the raw point cloud, producing a topologically consistent mesh with known
correspondences. This process is commonly referred as ``human body fitting''.

In particular, we focus on fitting the expressive parametric body model, \ie \smplx~\cite{SMPLX:2019},
which includes detailed hand gestures and facial expressions, to raw point clouds
of clothed humans. This is a challenging problem due to the large variation in
clothing styles, body shapes, and poses, as well as the presence of noise and partial
observations in the input point clouds.
Even worse, the 3D clothed scans with perfectly aligned \smplx ground-truth are extremely
scarce, and collecting such data is labor-intensive and costly.

Body fitting pipelines typically involve two steps: 1) establishing
correspondences between the input clothed human point cloud and the body model
template, like \smpl~\cite{SMPL:2015}, via ICP (iterative closest point) or its
variants~\cite{chen1992icp,allen2003space,pons2015dyna,zuffi2015stitched,feng2023arteq},
and 2) optimizing~\cite{bhatnagar2020loopreg,wang2021ptf,marin24nicp,easymocap,federica2016smplify,li2025etch,bhatnagar2020ipnet}
or regressing~\cite{feng2023arteq} the model parameters to minimize the distance
between corresponding points. The correspondence could be dense~\cite{bhatnagar2020loopreg,wang2021ptf,marin24nicp}
or sparse, such as inner keypoints~\cite{easymocap,federica2016smplify}, surface
markers~\cite{li2025etch}, and part-based labels~\cite{bhatnagar2020ipnet}.

Dense correspondence is inherently ill-defined for clothed humans, as the outer clothing
surface can deviate substantially from the underlying body, especially for loose
or dynamic garments. This deviation introduces ambiguity and instability, since
a single point on a loose T-shirt, for example, may correspond to multiple
possible locations on the inner body, and these correspondences can change as the
clothing deforms. While some recent works attempt to learn dense correspondences
to align the parametric body model to the clothed surface~\cite{bhatnagar2020loopreg,wang2021ptf,marin24nicp},
the resulting fitted bodies often appear unnatural, overweight, or biomechanically
implausible, as {illustrated in~\cref{fig:teaser}-A.}

For sparse correspondence, several practical solutions exist. For instance, 2D
keypoint estimators~\cite{fang2022alphapose,cao2019openpose} trained on in-the-wild
images of clothed humans can provide reasonably accurate and robust 2D keypoints,
even when the underlying body is occluded. Additionally, surface-based sparse markers
can be weighted and aggregated (voting with confidence) from dense correspondence~\cite{li2025etch}
to improve stability.
However, inner keypoints capture only the skeleton, providing limited
information about body shape (fat or slim).
Surface markers are often too sparse to capture fine details such as hand gestures
and facial expressions, {as illustrated in~\cref{fig:teaser}-B}, which are crucial
for many applications, like human-object interaction. Moreover, with partial
input point clouds, these sparse correspondences may be entirely absent, causing
the fitting process to fail, {see~\cref{fig:teaser}-C}.

The limitations of sparse correspondence -- namely, reduced expressiveness for fine-grained
parts and vulnerability to partial inputs -- can be mitigated by adopting dense
correspondence. However, as discussed above, dense correspondence becomes inherently
ambiguous in the presence of clothing. This raises a key question: how can we balance
\textbf{local expressiveness} and \textbf{global robustness} in human body fitting?
-- \textit{Undress first, then dense fit!}

However, this requires an accurate ``undressing'' operation, which is itself a challenging
problem.
Inspired by ETCH~\cite{li2025etch}, which learns \emph{SE(3) equivariant tightness
vectors} to effectively disentangle clothing from the underlying body, we extend
it as \modelname, by retaining the original tightness vector regressor while
replacing its explicit sparse markers with implicit full-body dense
correspondence~\cite{corona2022lvd,marin24nicp}.
The tightness vector regressor is critical for robust undressing, as the learned
tightness vectors are locally SE(3) equivariant, providing reliable cues even
when portions of the input point cloud are missing. The implicit full-body correspondence,
defined on the expressive \smplx~\cite{SMPLX:2019} model, enables dense matching
between undressed human scans and the body, capturing fine details such as hand
gestures and facial features. Furthermore, the implicit representation is
inherently robust to partial inputs -- after training with partial augmentation,
\textit{fullset} \smplx anchor points can be queried at any location in the entire
feature space.
\modelname, therefore, achieves robustness to partial inputs, and expressiveness
for fine details, {see~\cref{fig:teaser}-D}.

More importantly, such \textit{``undress first, then dense fit''} paradigm echos
the emerging trend of scaling efforts in computer vision~\cite{dosovitskiy2020vit,
wang2025vggt, siméoni2025dinov3}.
Since the ``undress'' and ``dense fit'' modules are disentangled, we can
independently leverage 1) unlimited simulated garments (\ie, \clothd~\cite{bertiche2020cloth3d})
and 2) large-scale pose libraries (\ie, \amass~\cite{mahmood2019amass} for body poses and \interhand~\cite{Moon_2020_ECCV_InterHand2.6M} for hand poses) to train
each module separately. This modular approach enables us to combine them for superior
robustness in fitting in-the-wild clothed human scans. In other words, simulated
garments enrich the diversity of clothing styles for tightness-aware undressing,
while large-scale pose libraries enhance the generalization of implicit dense
fitting to various bodies and hand gestures.

We conduct a comprehensive evaluation of \modelname against SOTA methods on both
in-distribution datasets, such as CAPE~\cite{ma2020cape} and 4D-Dress~\cite{wang20244ddress},
as well as out-of-distribution data from BEDLAM2.0~\cite{tesch2025bedlam2}. \modelname consistently
demonstrates superior performance in terms of expressiveness (\eg, accurately capturing
hand gestures and facial details) and robustness (\eg, effectively handling
partial inputs). To validate the effectiveness of the \textit{``undress first, then
dense fit''} paradigm, we benchmark against \textit{``dense fit only''} approaches
like NICP~\cite{marin24nicp} and \textit{``undress first, then sparse fit''}
methods such as ETCH~\cite{li2025etch}. Additionally, we ablate two technical
innovations: hand refinement by re-sampling, which produce better hand poses, and skin-aware tightness masking, which rectifies tightness vectors
on skin regions to improve undressing performance.
Finally, we conduct a scaling analysis on simulated garments, \clothd, and body pose
libraries, \amass, highlighting the potential for future scaling towards truly
generalizable human body fitting.

In summary, we upgrade ETCH~\cite{li2025etch} in three key aspects, with all
\textcolor{GreenColor}{$\%$} values indicating reduced error over ETCH:

\begin{itemize}[leftmargin=*,nosep]

    \item \textbf{More Expressive.} Replacing \smpl with \smplx, employing implicit
        dense correspondence, and introducing re-sampling based hand refinement, \modelname
        captures finer hands (V2V-\textcolor{GreenColor}{$35.8\%\downarrow$} on CAPE)
        and head (V2V-\textcolor{GreenColor}{$8.1\%\downarrow$} on \ddress),
        which are crucial for contact-rich interactions.

    \item \textbf{More Robust.} By decoupling the \textit{``undress''} and \textit{``dense
        fit''} modules, \modelname is robust to diverse clothing styles and pose
        variations. The locality of the tightness vector, replacement of sparse markers
        with implicit dense correspondence, and partial augmentation further
        improve its robustness on partial inputs (V2V-\textcolor{GreenColor}{$72.
        5\%\downarrow$} on \ddress) and unseen human captures (MPJPE-\textcolor{GreenColor}{$8
        0.8\%\downarrow$} on \bedlam).

    \item \textbf{More Scalable.} \modelname seamlessly scales with large-scale
        3D garments and pose libraries, both of which are easier to simulate or collect
        than real scans. This scalability further reduces fitting error (MPJPE-\textcolor{GreenColor}{$2
        7.2\%\downarrow$} on \bedlam). These results underscore the
        effectiveness of our modular design and highlight the potential for
        future scaling towards truly generalizable human body fitting.
\end{itemize}

    \section{Related Work}
\label{sec:related}

Fitting human body models to point clouds is fundamental to many human-centric tasks.
Over the years, a wide range of methods have been proposed to tackle this challenge.
We analyze  them
from three key perspectives: optimization vs.
learning, tightness-agnostic vs. tightness-aware, and sparse vs. dense
correspondence. We also clarify how \modelname is positioned within this landscape.

\pheading{Optimization vs. Learning.}
Early optimization-based human body fitting methods typically rely on the ICP
algorithm~\cite{chen1992icp} or its variants~\cite{allen2003space,pons2015dyna,zuffi2015stitched}.
Modern optimization-based approaches~\cite{zhang2017buff,ma2020cape,patel2021agora,zheng2019deephuman,tao2021function4d,rbh_reg,easymocap}
often involve complex pipelines with multiple intermediate steps, such as pose
estimation~\cite{cao2019openpose,fang2022alphapose}, body segmentation~\cite{antic2024close,Gong2019Graphonomy},
and triangulation, each potentially introducing errors that can accumulate and degrade
final accuracy. While optimization-based methods can achieve highly accurate
results given precise correspondences, they are generally time-consuming,
motivating the development of more efficient alternatives.

Learning-based methods leverage large-scale 3D human datasets~\cite{mahmood2019amass,wang20244ddress,ma2020cape}
and deep neural networks~\cite{qi2017pointnet,qi2017pointnet++,zaheer2017deep,thomas2019kpconv,zhao2021point,wu2022point,wu2024point}
designed for point cloud processing. These approaches either provide good initialization
for subsequent fitting~\cite{wang2021ptf,bhatnagar2020ipnet,bhatnagar2020loopreg},
directly regress body meshes~\cite{prokudin2019efficient,wang2020sequential,zhou2020reconstructing},
or predict statistical body model parameters~\cite{feng2023arteq,jiang2019skeleton,liu2021votehmr}
in a feed-forward manner, offering much faster inference but sometimes less
accuracy than optimization-based methods. To balance speed and accuracy, hybrid
approaches first estimate sparse markers~\cite{li2025etch} or dense
correspondences~\cite{marin24nicp}, and and then refine body parameters via optimization.
\modelname adopts this hybrid paradigm.

\pheading{Tightness-agnostic vs. Tightness-aware.}
Many methods~\cite{bhatnagar2020loopreg,feng2023arteq,marin24nicp},
optimization- or learning-based, fit the human body model directly to the input
point cloud. This works well for tight clothing, but fails for loose clothing, where
the true body shape can deviate significantly from the observed surface, leading
to inaccurate fits.
To address this, ``tightness-aware'' methods~\cite{chen2021tightcap, bhatnagar2020ipnet,
wang2021ptf, li2025etch}
explicitly model clothing to recover more accurate body shapes. TightCap~\cite{chen2021tightcap}
uses a clothing tightness field—displacements from garment to body in UV space.
IPNet~\cite{bhatnagar2020ipnet} and PTF~\cite{wang2021ptf} jointly predict inner
and outer body surfaces via double-layer occupancy. ETCH~\cite{li2025etch} encodes
tightness as displacement vectors from the cloth surface to the underlying body.
\modelname extends ETCH’s tightness vector by introducing skin-aware masking, setting
tightness to zero on uncovered skin regions.

\pheading{Sparse Correspondence vs. Dense Correspondence.}
Establishing correspondence is a crucial step in human body fitting, typically
categorized as either ``sparse'' or ``dense.'' Sparse correspondence~\cite{feng2023arteq,li2025etch}
often relies on part-based feature aggregation,
which provides some robustness to noise but 
struggle with incomplete input,
as missing regions may result in lost markers. In contrast, explicit dense correspondence~\cite{bhatnagar2020ipnet,bhatnagar2020loopreg,wang2021ptf,marin24nicp}
queries pointwise correspondence features from a learned implicit field.
Although dense sampling can be computationally intensive, it is generally more robust
to partial inputs—especially when combined with partial data augmentation—and better
captures fine details. \modelname adopts the implicit dense correspondence
strategy, following NICP~\cite{marin24nicp}, and extends it with
re-sampling based hand refinement to enhance hand pose accuracy.

    \section{Method}
\label{sec:method}

\begin{figure*}[t]
    \centering
    \includegraphics[width=\linewidth]{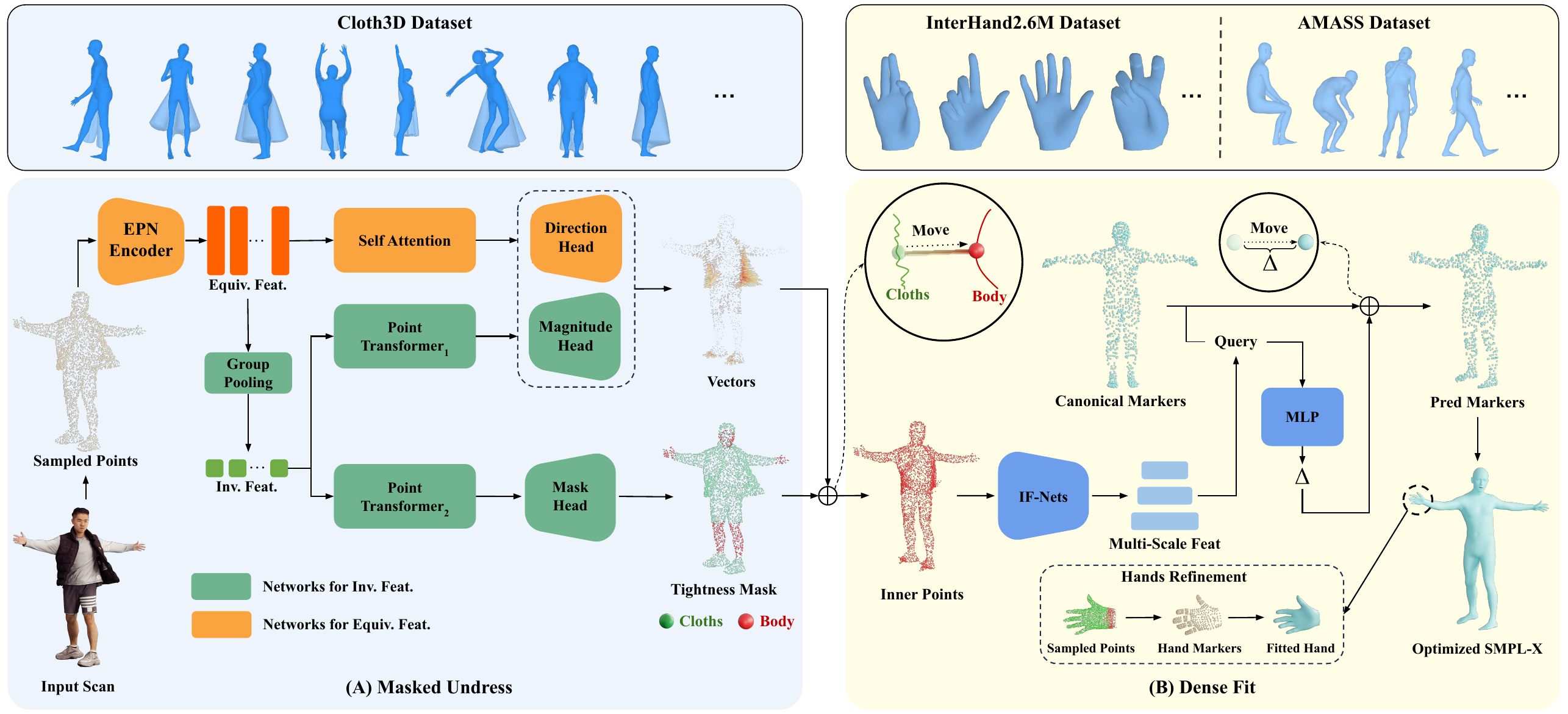}
    \vspace{-1.5em}
    \caption{\scriptsize \textbf{Two stages of \modelname: (A) Masked Undress, (B)
    Dense Fit}. In the Masked Undress stage, we take a clothed scan as input and
    compute the undressed body ($\hat{\mathbf{y}}_{i}= \mathbf{x}_{i}+ \hat{l}_{i}
    \hat{\mathbf{v}}_{i}$). In the Dense Fit stage, we implicitly learns the deforming
    field, which deforms the canonical \smplx into a posed one. Thanks to the decoupled
    design, the robustness to dynamic clothing and pose variations could be
    improved with simulated garments, \ie, \clothd~\cite{bertiche2020cloth3d}, and pose
    libraries, \ie, \amass~\cite{mahmood2019amass} for body poses and \interhand~\cite{Moon_2020_ECCV_InterHand2.6M} for hand poses, respectively.}
    \label{fig:pipeline}
    \vspace{-1.5em}
\end{figure*}

As illustrated in~\cref{fig:pipeline}, following the \textit{``undress first, then dense fit''} paradigm, our method \modelname consists of two stages: 1) \textit{masked undress}, which learns \emph{SE(3) equivariant tightness vectors} and \emph{skin mask} to obtain inner points from the clothed point cloud (\cref{subsec:undress}); 2) \textit{dense fit}, which encodes the inner points implicitly to establish dense correspondence for \smplx model fitting (\cref{subsec:dense_fit}).

Basically, \modelname extends ETCH~\cite{li2025etch} by replacing the \smpl model with the more expressive \smplx model~\cite{SMPLX:2019} and 
replacing the explicit sparse markers with implicit dense correspondence as in NICP~\cite{marin24nicp}, which is not only more robust to partial inputs,
but also enables more detailed fitting, particularly for the hands and face. Furthermore, the ``undress'' and ``dense fit'' modules are well disentangled and can be trained separately with garment-rich data (\ie, \clothd~\cite{bertiche2020cloth3d}) and pose-rich data (\ie, \amass~\cite{mahmood2019amass} and \interhand~\cite{Moon_2020_ECCV_InterHand2.6M}) for robust regression of clothing tightness and body/hand poses, making the training process more flexible and scalable.

\subsection{Masked Undress: Clothed to Body Points}
\label{subsec:undress}

\qheading{Tightness Vector~\cite{li2025etch}.} ETCH proposes to model cloth-to-body tightness using a set of vectors, \ie, tightness vectors. ``Tightness vector $\mathbf{v}_{i}$'' is the point-wise 3D vector pointing from the outer point $\mathbf{x}_{i}$ (clothed human body) to the inner point $\mathbf{y}_{i}$ (underneath body), \ie $\mathbf{y}_{i} = \mathbf{x}_{i} + \mathbf{v}_{i}$. The tightness vector comprises two components: direction $\mathbf{d}_{i}$ and magnitude $b_{i}$, \ie $\mathbf{v}_{i} = b_{i} \mathbf{d}_{i}$.
Given a point cloud  $\vect{X} = \{ \vect{x}_i \in \real^3 \}_N$ that is randomly sampled from the 3D clothed humans, ETCH uses an EPN~\cite{chen2021equivariant} to produce an $\mathrm{SO}(3)$-equivariant feature 
$\mathcal{F}$, mean pooling over this feature yields an invariant feature $\overline{\mathcal{F}_\text{EPN}}$, or abbreviated as $\overline{\mathcal{F}}$.

The direction highly correlates with human articulated poses, thus it is learned with local approximate SE(3) equivariant features, ensuring the tightness vector consistently maps the cloth surface to the body surface. Specifically, a self-attention network $\mathcal{F}_\text{self-attn}$ is used to process the equivariant feature $\mathcal{F}$ over the rotation group dimension ($O$), to ensure that each group element feature is associated with a group element $\mathbf{g}_j$, then a direction head  $\mathcal{F}_\text{Direc}$ parameterized with an MLP network followed by tranformations is used to process the output feature to help to produce the final direction $\hat{\mathbf{d}}_i$.
    While the magnitude mainly reflects clothing displacements, which highly correlate with clothing types and body regions, thus, it is learned from the invariant features. From the invariant feature $\overline{\mathcal{F}}$, a \emph{Point Transformer}~\cite{zhao2021point} $\mathcal{F}_\text{PT-1}$ is used to capture the contextual information outside each point, then a magnitude MLP head $\mathcal{F}_\text{Mag}$ produces $\hat{b}_i$. Finally, the tightness vectors are obtained via $\hat{\mathbf{v}}_{i} = \hat{b}_{i} \hat{\mathbf{d}}_{i}$:
    \begin{equation}
    \begin{aligned}
        \hat{\mathbf{d}}_{i} = \mathcal{F}_\text{Direc}(\mathcal{F}_\text{self-attn}(\mathcal{F}_\text{EPN}(\mathbf{X})_{i})), 
        \hat{b}_{i} = \mathcal{F}_\text{Mag}(\mathcal{F}_\text{PT-1}(\overline{\mathcal{F}}(\mathbf{X})_{i}, \mathbf{x}_{i}; \delta)).
    \end{aligned}
    \end{equation}
    where $\delta = \Theta(\mathbf{x}_{i} - \mathbf{x}_{j})$ is the learned position embedding to encode the relative positions between point pairs $\{\mathbf{x}_{i}, \mathbf{x}_{j}\}$.

\qheading{Tightness Masking.}
Since not all surface points exhibit non-zero tightness (\ie, regions such as the head, hands, and exposed skin), we introduce a tightness mask for more precise undressing. 
Determining whether a surface point should exhibit non-zero tightness is naturally a binary classification problem, \ie, we assign a label $l_i$ to each point, `1' for non-zero tightness and `0' for zero tightness. 
Inspired by ETCH, we use Point Transformer $\mathcal{F}_\text{PT-2}$ and $\mathcal{F}_\text{Label}$ takes $\overline{\mathcal{F}}(\mathbf{X})  \in \mathbb{R}^{N \times C}$ with position $\mathbf{X} \in \mathbb{R}^{N \times 3}$ as input, and
outputs $\mathcal{P} \in \mathbb{R}^{N \times 2}$,
represents the probability of a point $\mathbf{x}_i$ belonging to each class (zero vs. non-zero tightness):
\begin{equation}
    \begin{aligned}
     \mathcal{P}(\mathbf{X}) = \texttt{softmax}(\mathcal{F}_\text{Label}(\mathcal{F}_\text{PT-2}(\overline{\mathcal{F}}(\mathbf{X})_{i}, \mathbf{X}; \delta))),
     \hat{L} = \texttt{argmax} (\mathcal{P}(\mathbf{X})).
     \end{aligned}
\end{equation} 
\qheading{Training and Inference.}
We regress the direction $\hat{\mathbf{d}}_i$, the magnitude $\hat{b}_i$, and the label $\hat{l}_i$ for each point $\mathbf{x}_i$.
The final training loss $\mathcal{L}$ is formulated as follows:
\begin{equation}
\label{eq:losses}
    \begin{gathered}
        \mathcal{L} = w_d \mathcal{L}_d + w_b \mathcal{L}_b + w_l \mathcal{L}_l ,\\
        \mathcal{L}_d = \sum_{i=1}^{N} \hat{l}_i\frac{\hat{\mathbf{d}}_i \cdot \mathbf{d}_i}{\|\hat{\mathbf{d}}_i\| \|\mathbf{d}_i\|},  \mathcal{L}_b = \frac{1}{N} \hat{l}_i\sum_{i=1}^{N} (\hat{b}_i - b_i)^2,
        \mathcal{L}_l = - \frac{1}{N} \sum_{i=1}^{N} \log(\mathcal{P}(\mathbf{x}_i, l_i)).
    \end{gathered}
\end{equation}
During inference, we obtain the inner point $\hat{\mathbf{y}}_{i}$ by $\hat{\mathbf{y}}_{i} = \mathbf{x}_{i} + \hat{l}_i\hat{\mathbf{v}}_{i}$, where $\hat{\mathbf{v}}_{i} = \hat{b}_{i} \hat{\mathbf{d}}_{i}$ for each point $\mathbf{x}_{i}$. Finally we get the inner body point clouds $\hat{\vect{Y}} = \{ \hat{\vect{y}}_i \in \real^3 \}_N$ that will be used for the following dense fit stage.

\subsection{Dense Fit: Body Points to \smplx}
\label{subsec:dense_fit}

\qheading{\smplx~\cite{SMPLX:2019}} 
extends \smpl with fully articulated hands and an expressive face.
As a statistical body model, \smplx maps body shape $\shapecoeff \in \real^{10}$, facial expression $\expressioncoeff \in \real^{10}$ and pose $\posecoeff \in \real^{J \times 3}$ parameters to mesh vertices $\vect{V}\in \real^{10475 \times 3}$, where $J$ is the number of human joints ($J=55$, containing body, eyes, jaw and finger joints in addition to a joint for global rotation). $\shapecoeff$ are linear shape coefficients of the shape blend shape function, and $B_{S}(\shapecoeff)$ accounts for variations of body shapes. $\posecoeff$ contains the relative rotation (axis-angle) of each joint plus the root one \wrt their parent in the kinematic tree, and $B_{P}(\posecoeff)$ models the pose-dependent deformation. $\expressioncoeff$ are PCA coefficients of the expression blend shape function, and $B_{E}(\expressioncoeff)$ accounts for variations of facial expressions. Shape displacements $B_{S}(\shapecoeff)$, pose correctives $B_{P}(\posecoeff)$ and facial expression displacements $B_{E}(\expressioncoeff)$ are added together onto the template mesh $\template \in \real^{10475 \times 3}$, in the rest pose (or T-pose), to produce the output mesh $\restpose$:
\begin{equation}
    \label{eq:smplx_t}\restpose(\shapecoeff, \posecoeff, \expressioncoeff) = \template
    + B_{S}(\shapecoeff) + B_{P}(\posecoeff)+B_{E}(\expressioncoeff),
\end{equation}
Next, the joint regressor $J(\shapecoeff)$ is applied to the rest-pose mesh $\restpose$ to obtain the 3D joints : $\real^{\shapedim}\to \real^{J \times 3}$. Finally, Linear Blend Skinning (LBS) $W(\cdot)$ is used for reposing purposes, the skinning weights are denoted as $\blendweights$, then the posed mesh is translated with $\boldsymbol{t}\in \real^{3}$ as final output $\mathbf{M}$ :
\begin{equation}
    \label{eq:smplx_lbs}\mathbf{M}(\shapecoeff, \posecoeff, \expressioncoeff, \mathbf{t}
    ) = W( \restpose(\shapecoeff, \posecoeff, \expressioncoeff), J(\shapecoeff),
    \posecoeff, \boldsymbol{t}, \blendweights ). %
\end{equation}

\qheading{Neural ICP (NICP)~\cite{marin24nicp}}
is a test-time tuning method for human body fitting.
Inspired by LVD~\cite{corona2022lvd}, given an inner point cloud $\mathbf{Y}$, NICP first uses IF-Nets~\cite{chibane2020implicit} $\mathcal{F}_{\text{IF}}$ to encode it into implicit feature volume $\mathbf{Z}$. Then it learns a neural field $\mathcal{F}_{\text{NF}}$, which is represented by an MLP, given any query point $\mathbf{q} \in \real^3$, the neural field predict the ordered offsets from the query point to the target \smplx body model vertices:
\begin{equation}
    \begin{aligned}
        \mathbf{Z}= \mathcal{F}_{\text{IF}}(\mathbf{Y}),\quad \mathbf{o}= \mathcal{F}_{\text{NF}}(\mathbf{Z}(\mathbf{q}))
    \end{aligned}
\end{equation}
where $\mathbf{Z}(\mathbf{q})$ denotes the feature queried from the feature volume $\mathbf{Z}$ from position $\mathbf{q}$, and $\mathbf{o} \in \real^{N}$ denotes the offsets from the position $\mathbf{q}$ to a subset of target SMPL vertices. 

Unlike LVD, which predicts offsets to all template vertices (e.g., 10475 for \smplx) using a single MLP, NICP introduces LoVD: a local variant with multiple MLP heads, each specialized for a body region (16 regions via spectral clustering). The template vertices are also downsampled by a factor of 10 (e.g., 1051 for SMPL) for efficiency.

NICP preforms test-time fine-tuning for better robustness. Specifically, after obtaining the neural field, NICP fine-tunes it on the input inner point cloud $\mathbf{Y}$ by iterative optimization. First, NICP samples $\mathbf{y}_{k}$ from $\mathbf{Y}$ as query points, and find its correspondence vertex ID $i_{k}$ on the SMPL template by
\begin{equation}
    \begin{aligned}
        i_{k} = \arg\min_{i} \|\mathcal{F}_{\text{NF}}(\mathbf{Z}(\mathbf{y}_{k}))_{i}\|^{2}_{2}
    \end{aligned}
\end{equation}

\begin{figure}[t]
    \centering
    \includegraphics[width=\linewidth]{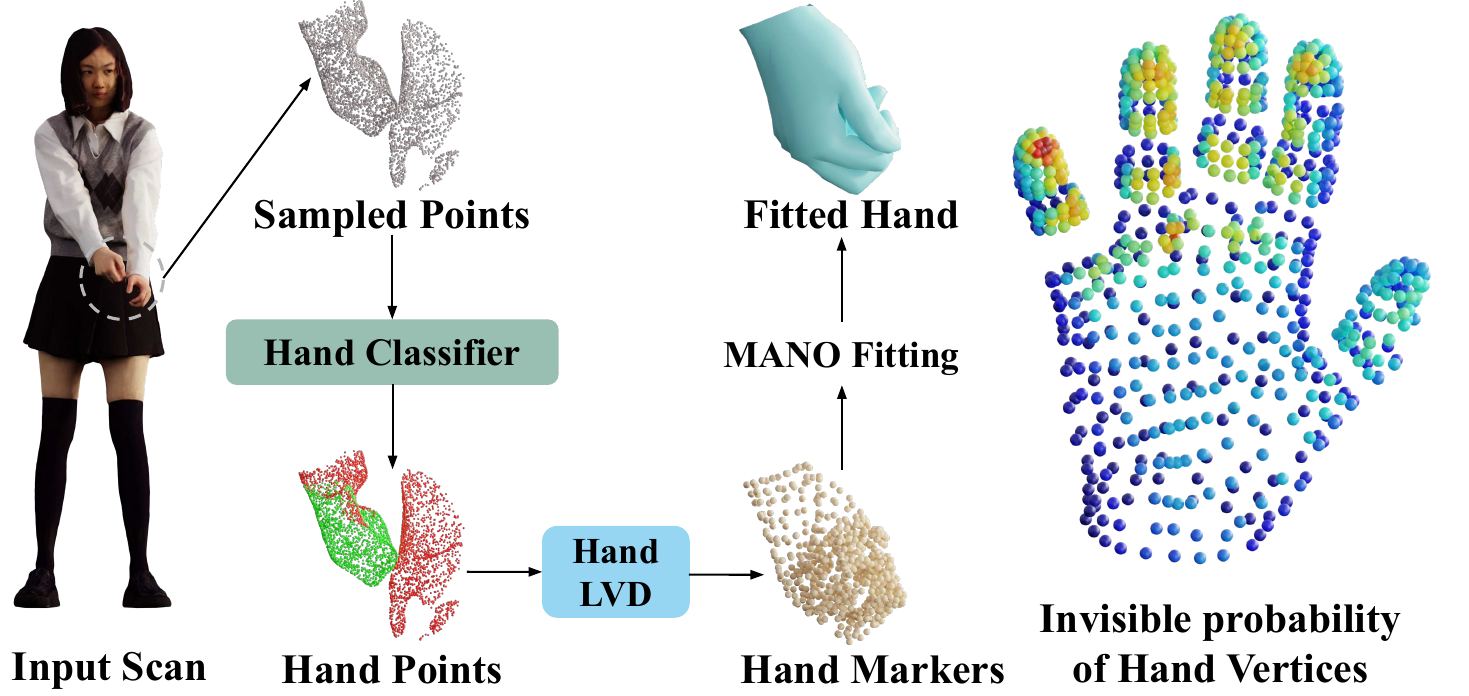}
    \caption{\scriptsize \textbf{Hand Refinement by Re-sampling.} After obtaining initial body fitting, we re-sample points around the hand and fit hand model separately.}
    \label{fig:hand_refine}
\end{figure}

Then it minimizes the distance between correspondence points by updating the parameters $\theta$ of the neural field:

\begin{equation}
    \begin{aligned}
        \theta^{*} = \arg\min_{\theta} \sum^{n}_{k=1}\|\mathcal{F}_{\text{NF}}(\mathbf{Z}(\mathbf{y}_{k}))_{i_k}\|^{2}_{2}
    \end{aligned}
\end{equation}
The test-time fine-tuning helps improve performance and robustness as depicted in NICP~\cite{marin24nicp}. Then the target SMPL vertices ${\{\mathbf{m}_j\}}_J$ are obtained by querying the neural field for model fitting. When $J$ vertices  are used, the SMPL -X model parameters are fitting by minimizing:
\begin{equation}
    \min_{\boldsymbol{\theta}, \boldsymbol{\beta},\expressioncoeff,\mathbf{t}} \sum_{j=1}^{J} \left\| \mathbf{m}_j - \mathbf{M}(\shapecoeff, \posecoeff,\expressioncoeff,  \mathbf{t}
    ) _j \right\|^2_2
\end{equation}
\qheading{Hand Refinement by Re-sampling.} 
Although the hand pose can be fitted while fitting the \smplx body, the hand pose is often inaccurate because the sampling points for the hand are usually sparse. To improve hand pose, inspired by some image-based human reconstruction methods that detect and reconstruct hands separately in images~\cite{cai2023smplerx}, we adopted a hand refinement strategy based on re-sampling, as shown in~\cref{fig:hand_refine}. 

Specifically, after obtaining the initial fitted body, we know the approximate position of the hand. Based on this, we resample near the initial hand position to obtain denser hand sampling points. We then individually fit the hand model, MANO~\cite{mano}, to these sampling points. However, as TUCH~\cite{tuch} has shown, the hand easily comes into contact with other body parts, causing the sampled points to include parts that do not belong to the hand, and these points can affect the hand fit results. Thus, we train a hand classifier on MTP dataset proposed in~\cite{tuch} to remove points that do not belong to the target hand. Then the hand points can be used by a hand LVD network, similar to~\cite{corona2022lvd, marin24nicp} to produce hand markers, which are finally fitted to MANO hand model.
As shown in the right of~\cref{fig:hand_refine}, due to self-contact and occlusion, many vertices of the hand may be invisible. Therefore, when training the hand LVD, we augment the hand data based on the probability distribution of invisible hand vertices calculated from the MTP dataset. These design strategies are validated in
ablation studies~\cref{subsec:exp_ablation}.

\qheading{Training and Inference.}
We follow the same training scheme as NICP~\cite{marin24nicp} to train the LoVD. The neural field predicts 1051 offsets for each query points. At test time, we also perform iterative optimization to fine-tune the LoVD, then obtain the dense correspondence by querying the neural field for following \smplx model fitting. 
After obtaining the fitted \smplx body, we perform hand refinement, the fitted MANO hands are then transformed back to the \smplx model.

    \section{Experiments}
\label{sec:experiments}

\subsection{Datasets}
\label{subsec:exp_datasets}

For fair and comprehensive assessment, we follow established benchmarks~\cite{wang2021ptf,li2025etch,feng2023arteq,bhatnagar2020ipnet,marin24nicp}
and evaluate our method on CAPE~\cite{ma2020cape} and \ddress~\cite{wang20244ddress}.
Given \modelname's disentangled \textit{undress} and \textit{dense fit} modules,
we train them separately on garment-rich (\clothd~\cite{bertiche2020cloth3d})
and pose-rich (\amass~\cite{mahmood2019amass}) datasets to analyze scalability.
Our error analysis shows that CAPE and \ddress fitted bodies can be inaccurate,
potentially affecting evaluation. To mitigate this, we also benchmark all
methods on \bedlam~\cite{tesch2025bedlam2}, which provides simulated clothing with
perfect body fits. As no model is trained on \bedlam, it serves as a pure out-of-distribution
(OOD) test, better reflecting generalization. Finally, we describe how we
simulate partial scans.

\pheading{CAPE~\cite{ma2020cape} } contains 15 subjects with different body shapes;
we split them as 4:1 to evaluate the robustness against \textit{various body
shapes and garments}. As NICP~\cite{marin24nicp}, we subsample by factors of 5 and
20 for the training and validation sets, resulting in 26,004 train frames and 1,021
valid frames.

\qheading{\ddress~\cite{wang20244ddress} } has loose clothing and a large range of
motion; it contains 32 subjects with 64 outfits across over 520 motions.
We use the official split, which selects 2 sequences per outfit, to evaluate the
robustness against ``body pose and clothing dynamics variations''. After
subsampling by factors of 1 and 10 for training and validation, we obtain 59,395
train frames and 1,943 valid frames.

\qheading{\clothd~\cite{bertiche2020cloth3d} } is a large-scale simulated
dataset of 3D clothed human. It contains a large variability in garment type,
topology, shape, size, tightness, and fabric. Dynamic clothes are simulated on top
of thousands of different pose sequences and body shapes. It contains more than
2M frames (8K+ sequences) of simulated and rendered garments in 7 categories. We
downsampled the full dataset and built roughly 150k paired simulated 3D human
scans.

\qheading{\amass~\cite{mahmood2019amass} } is a large motion database that unifies
different optical marker-based mocap datasets. It contains more than 11,000
motions, covering a wide range of scenarios. As NICP~\cite{marin24nicp}, we adopt
the official splits and obtain a trainset with roughly 120k \smplx bodies by downsampling.

\qheading{\bedlam~\cite{tesch2025bedlam2} } is a large-scale synthetic video
dataset of animated bodies in simulated clothing, containing more than 8M images.
it is a significant expansion of the \bedlam dataset~\cite{tesch2025bedlam2}, which
increases pose and body BMI variation. It provides complete render assets, including
body textures, clothing assets, and so on. We randomly sampled 20 subjects with
various clothing from the dataset, each with 50 poses, totaling 1,000 paired
simulated 3D human scans.

\qheading{MTF~\cite{tuch}} dataset has 3731 images from 148 different subjects, mimicking poses with self-contact sampled from 3DCP Scan, 3DCP Mocap and AGORA. We use MTF data for training hand classifier and computing  the probability distribution of invisible hand vertices data augmentation. We also use \textbf{InterHand2.6M~\cite{Moon_2020_ECCV_InterHand2.6M}} dataset to train hand LVD.

\qheading{Partial Data.} We simulate the most common partial data pattern,
single-view, \ie, from a certain view angle, only the front part is visible.
Given a full mesh, to simulate the single-view mesh, we need to calculate the
intersection point of a rays emitted from a specific angle with the mesh surface.
We implement this using the Embree~\cite{embree} library.

\subsection{Full-scan Comparison}
\label{subsec:exp_full}

\begin{table*}
    [t]
    \centering
    \caption{\scriptsize \textbf{In-distribution Quantitative Comparison with SOTAs.} \modelname
    clearly outperforms SOTAs, whether tightness-agnostic (A.) or -aware (B.),
    in both CAPE and \ddress across almost all metrics. In 4D-Dress-V2V, it
    surpasses the ETCH by nearly \textcolor{GreenColor}{$21.2\%$}. 
    }
    \vspace{-1.0 em}
    \renewcommand{\arraystretch}{1.2}
    \resizebox{\textwidth}{!}{
    \begin{tabular}{c|c|c|cccc|cccc|c|cccc|cccc}
        \bottomrule
        \multirow{3}{*}{Groups}     & \multirow{3}{*}{Methods} & \multicolumn{9}{c|}{CAPE} & \multicolumn{9}{c}{4D-Dress}           \\
        \cline{3-20}                &                          & CD $\downarrow$           & \multicolumn{4}{c|}{V2V $\downarrow$} & \multicolumn{4}{c|}{MPJPE $\downarrow$} & CD $\downarrow$ & \multicolumn{4}{c|}{V2V $\downarrow$} & \multicolumn{4}{c}{MPJPE $\downarrow$} \\
        \cline{3-20}                &                          & All                       & All                                   & Hands                                   & Head            & Other                                 & All                                   & Hands          & Head           & Other          & All            & All            & Hands          & Head           & Other          & All            & Hands          & Head           & Other          \\
        \Xhline{0.8pt} \multirow{2}{*}{A.} & NICP                     & -                         & 1.736                                 & \second{2.741}                          & \second{1.184}  & 1.827                                 & 2.074                                 & \second{2.565} & 1.042          & 1.597          & -              & 4.085          & 6.224          & 3.323          & 3.993          & 4.862          & 6.142          & 2.540          & 3.521          \\
                                    & ArtEq                    & -                         & 2.202                                 & 3.417                                   & 2.011           & 1.943                                 & 2.405                                 & 3.055          & 1.693          & 1.589          & -              & 3.072          & \second{4.537} & 3.145          & 2.636          & \second{3.378} & \second{4.170} & 2.335          & 2.156          \\
        \hline
        \multirow{4}{*}{B.}         & IPNet                    & 1.077                     & 5.529                                 & 7.454                                   & 5.485           & 5.001                                 & 5.611                                 & 6.600          & 4.527          & 4.399          & 1.187          & 7.495          & 8.881          & 7.378          & 7.178          & 7.380          & 8.606          & 5.973          & 5.894          \\
                                    & PTF                      & 1.194                     & 2.341                                 & 3.880                                   & 2.038           & 2.099                                 & 2.641                                 & 3.377          & 1.720          & 1.767          & 1.207          & 3.297          & 4.938          & 3.338          & 2.785          & 3.567          & 4.607          & 2.612          & 2.248          \\
                                    & ETCH                     & \second{1.040}            & \second{1.567}                        & 3.449                                   & 1.236           & \best{1.240}                          & \second{2.002}                        & 2.833          & \best{0.928}   & \best{1.007}   & \second{1.134} & \second{2.408} & 5.108          & \second{1.997} & \second{2.178} & 3.459          & 4.695          & \second{1.420} & \second{2.141} \\
                                    & ETCH-X                   & \best{1.015}              & \best{1.484}                          & \best{2.215}                            & \best{1.120}    & \second{1.341}                        & \best{1.764}                          & \best{2.148}   & \second{0.969} & \second{1.215} & \best{1.060}   & \best{1.897}   & \best{3.101}   & \best{1.836}   & \best{1.681}   & \best{2.317}   & \best{3.065}   & \best{1.391}   & \best{1.454}   \\
        \toprule
    \end{tabular}
    }
    
    \label{tab:comparison_combined}
\end{table*}

\begin{figure*}[t]
    \centering
    \includegraphics[width=\linewidth]{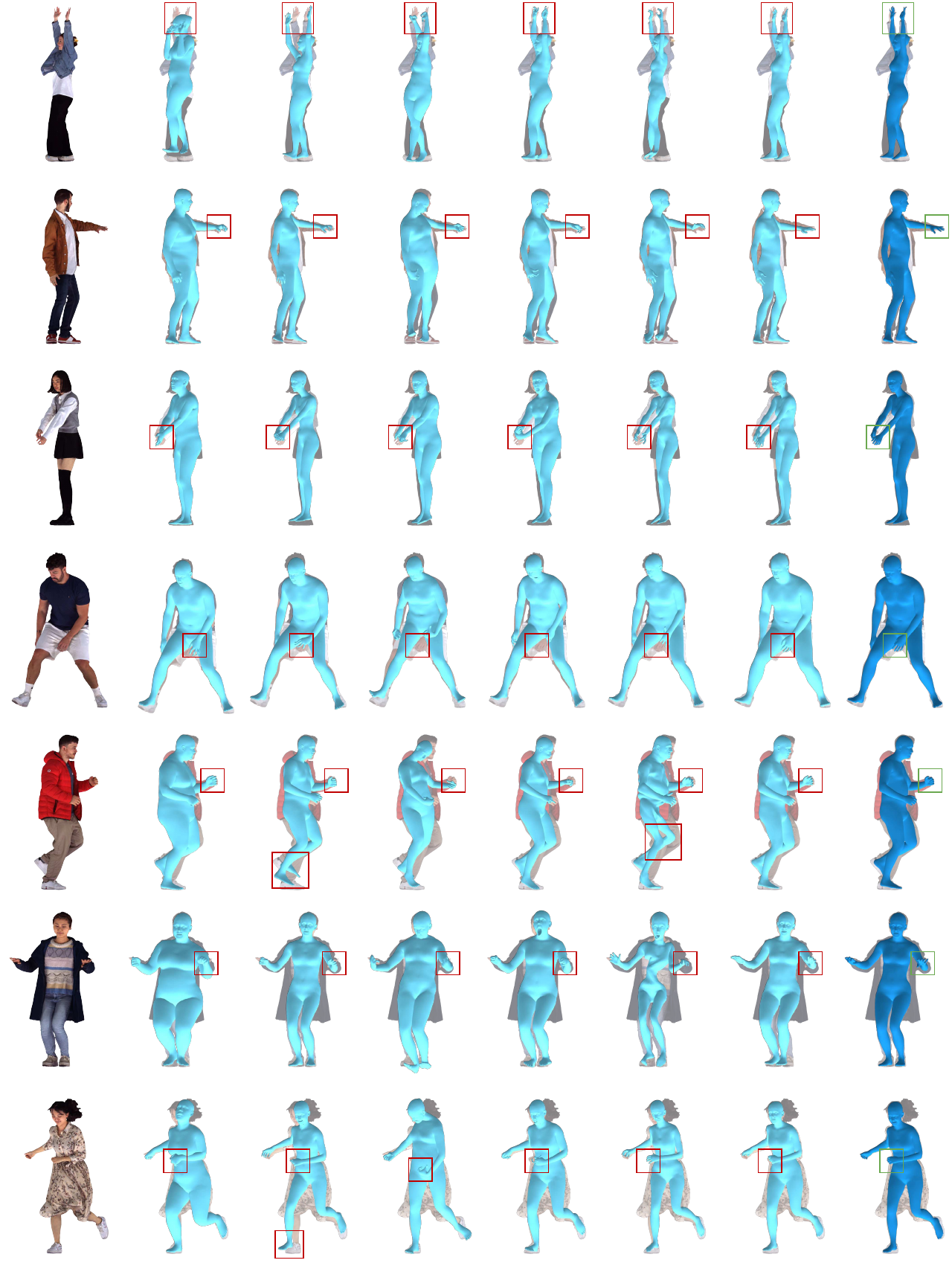}
    \begin{tabularx}
        {\linewidth}{ @{}p{0.020\linewidth}@{} >{\centering\arraybackslash}p{0.118\linewidth} >{\centering\arraybackslash}p{0.118\linewidth} >{\centering\arraybackslash}p{0.118\linewidth} >{\centering\arraybackslash}p{0.118\linewidth} >{\centering\arraybackslash}p{0.118\linewidth} >{\centering\arraybackslash}p{0.118\linewidth} >{\centering\arraybackslash}p{0.118\linewidth} >{\centering\arraybackslash}p{0.118\linewidth} }
        & \scriptsize Scan & \scriptsize \nicp~\cite{marin24nicp} & \scriptsize
        \arteq~\cite{feng2023arteq} & \scriptsize \ipnet~\cite{bhatnagar2020ipnet}
        & \scriptsize \ptf~\cite{wang2021ptf} & \scriptsize \etch~\cite{li2025etch}
        & \scriptsize \modelname & \scriptsize GT\\
    \end{tabularx}
    \caption{\scriptsize \textbf{Comparison with SOTAs on 4D-Dress.}}
    \label{fig:comparison}
\end{figure*}

We compare our method, \modelname, with multiple \sota baselines~\cite{bhatnagar2020ipnet,
wang2021ptf, marin24nicp,li2025etch}, as shown in \cref{tab:comparison_combined},
and the qualitative visualization comparison results on \ddress are shown in \cref{fig:comparison}.
Note that \modelname predicts SMPL-X bodies while previous methods only predict
SMPL bodies, for the fair of comparison, we implement the SMPL-X version of the
methods, and all results are calculated based on the SMPL-X body model.
\begin{table}[t]
    \centering
    \begin{minipage}[t]{0.35\textwidth}
    \centering
    \caption{\scriptsize \textbf{OOD Evaluation}. Note that all methods are trained on \ddress and test on BEDLAM2.0. }
    \vspace{-1.0 em}
    \renewcommand{\arraystretch}{1.2}
    \resizebox{0.96\linewidth}{!}{
    \begin{tabular}{c|c|ccc}
    \bottomrule
    Methods  & CD $\downarrow$ & V2V $\downarrow$ & MPJPE $\downarrow$ \\ \Xhline{0.8pt} \nicp &- &5.178 &6.238\\ \arteq  &- &4.136 &4.447\\ \ipnet  & 1.369& 8.641&9.471\\ \ptf  &1.288 &3.974 & 4.668\\ \etch  & 1.454 & 12.209 & 15.031 \\ \hline ETCH-X & \textbf{1.265} & \textbf{3.429} & \textbf{4.033} \\ 
    \toprule
    \end{tabular}}
    
    \label{tab:comparison_bedlam}
    \end{minipage}
    \hfill%
    \begin{minipage}[t]{0.63\textwidth}
    \centering
    \caption{\scriptsize \textbf{Evaluation Results for Partial Input}. Notably, single direction chamfer
  distance (CD) is used here.}
  \vspace{-1.0 em}
  \renewcommand{\arraystretch}{1.2}

  \resizebox{\linewidth}{!}{
  \begin{tabular}{c|c|ccc|ccc}
    \bottomrule
    \multirow{2}{*}{Train} & \multirow{2}{*}{Test}    & \multicolumn{3}{c|}{CAPE}        & \multicolumn{3}{c}{4D-Dress}      \\
                           &                          & CD $\downarrow$                  & V2V $\downarrow$                 & MPJPE $\downarrow$               & CD $\downarrow$                  & V2V $\downarrow$                 & MPJPE $\downarrow$               \\
    \Xhline{0.8pt} w/o Aug        & \multirow{2}{*}{Full}    & \textbf{0.894}                   & \textbf{1.484}                   & \textbf{1.764}                   & 0.951                            & \textbf{1.897}                   & \textbf{2.317}                   \\
    w/ Aug                 &                          & 0.918                            & 1.644                            & 2.027                            & \textbf{0.917}                   & 2.135                            & 2.677                            \\
    $\Delta$               &                          & \textcolor{RedColor}{$2.7\%$}    & \textcolor{RedColor}{$10.8\%$}    & \textcolor{RedColor}{$14.9\%$}    & \textcolor{GreenColor}{$3.6\%$}  & \textcolor{RedColor}{$12.5\%$}    & \textcolor{RedColor}{$15.5\%$}    \\
    \hline
    w/o Aug                & \multirow{2}{*}{Partial} & 1.149                            & 10.056                           & 10.403                           & 2.261                            & 13.861                           & 16.662                           \\
    w/ Aug                 &                          & \textbf{0.951}                   & \textbf{2.898}                   & \textbf{3.516}                   & \textbf{0.978}                   & \textbf{3.808}                   & \textbf{5.273}                   \\
    $\Delta$               &                          & \textcolor{GreenColor}{$17.2\%$} & \textcolor{GreenColor}{$71.2\%$} & \textcolor{GreenColor}{$66.2\%$} & \textcolor{GreenColor}{$56.7\%$} & \textcolor{GreenColor}{$72.5\%$} & \textcolor{GreenColor}{$68.4\%$} \\ %
    \toprule
  \end{tabular}
  }
  
  \label{tab:ablation_partial}
    \end{minipage}%
    
    \vspace{-1em}

\end{table}

\begin{figure}[t]
    \centering
    \includegraphics[width=\linewidth]{
        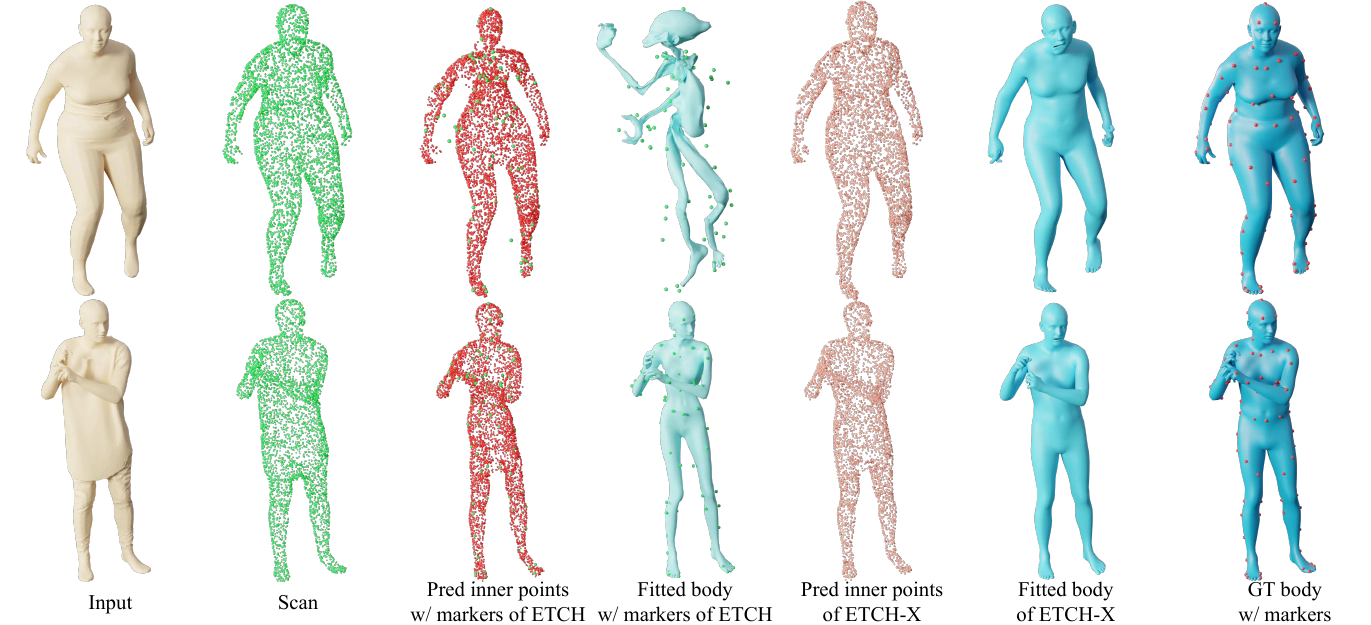
    }

    \vspace{-0.4 em}
    \caption{\textbf{Failure Case of ETCH~\cite{li2025etch} on BEDLAM2.0}.
    Two respresentative reasons for ETCH failure are incorrect part labeling (above) and inaccurate inner points (both). The failure is
    reflected in the large V2V (12.209cm) and MPJPE (15.031cm) errors of ETCH
    reported in \cref{tab:comparison_bedlam}}
    \label{fig:etch_bedlam2_failure}
\end{figure}

Overall, \modelname achieves superior performance across all datasets and
metrics. In particular, on CAPE, among all the competitors, our approach reduces
the V2V error by \textcolor{GreenColor}{$5.3\% \sim 73.2\%$} and MPJPE by \textcolor{GreenColor}{$11.9\% \sim 68.6\%$}; on 4D-Dress, the improvement is even more significant with a
\textcolor{GreenColor}{$21.2\% \sim 74.7\%$} decrease in V2V error and \textcolor{GreenColor}{$31.4\% \sim 68.6\%$} in MPJPE.
Among tightness-aware methods (\ie, \ipnet, PTF, ETCH and Ours), under bidirectional
Chamfer Distance, our method achieves \textcolor{GreenColor}{$2.4\% \sim 16.4\%$}
improvement on CAPE and \textcolor{GreenColor}{$7.0\% \sim 14.3\%$} on \ddress
between the predicted inner points/meshes (w/o SMPL-X fitting) and \gt SMPL-X
bodies.

Beyond in-distribution evaluation, we also assess out-of-distribution (OOD)
generalization in~\cref{tab:comparison_combined} on our \bedlam test set, with all
methods trained on the 4D-Dress for fairness. \modelname demonstrates notably
stronger generalization, achieving \textcolor{GreenColor}{$1.8\% \sim 13.0\%$}
lower Chamfer Distance, \textcolor{GreenColor}{$13.7\% \sim 71.9\%$} lower V2V, and
\textcolor{GreenColor}{$9 .3\% \sim 73.2\%$} lower MPJPE. ETCH, in particular, performs
poorly on V2V and MPJPE, likely due to limited generalization caused by  its entangled architecture design, as illustrated in \cref{fig:etch_bedlam2_failure}.

\subsection{Ablation Studies}
\label{subsec:exp_ablation}

\pheading{Partial Input.}
\begin{table}[t]
    \centering
    \hfill%
    \begin{minipage}[t]{0.46\textwidth}
    \centering
    \caption{\scriptsize \textbf{The Disentangled Design Enables Various Data Sources}.}
    \vspace{-1em}
  \renewcommand{\arraystretch}{1.2}

  \resizebox{\linewidth}{!}{
  \begin{tabular}{c|c|ccc|ccc}
    \bottomrule
    \multirow{2}{*}{Methods} & \multirow{2}{*}{Train Data} & \multicolumn{3}{c|}{CAPE} & \multicolumn{3}{c}{4D-Dress} \\
    \cline{3-8}          
    &    & CD $\downarrow$           & V2V $\downarrow$            & MPJPE $\downarrow$ & CD $\downarrow$ & V2V $\downarrow$ & MPJPE $\downarrow$ \\
    \Xhline{0.8pt} NICP     &   \amass      & -                         & {2.029}                     & {2.438}            & -               & {5.416}          & {5.280}            \\
    ETCH            &     \clothd     & {1.182}                   & 2.465                       & 2.993              & {1.560}         & 6.989            & 7.356              \\
    ETCH-X            &   \clothd+\amass     & \textbf{1.174}              & \textbf{1.975}                & \textbf{2.365}       & \textbf{1.515}    & \textbf{4.256}     & \textbf{4.200}       \\
    \toprule
  \end{tabular}
  }
  
  \label{tab:gen_data_comparison}
        \end{minipage}
        \hfill%
    \begin{minipage}[t]{0.495\textwidth}
    \centering
    \caption{\scriptsize \textbf{Ablation Study of Tightness Masking}. The models are trained on CLOTH3D+AMASS. 
    }
    \vspace{-1em}
    \renewcommand{\arraystretch}{1.2}

    \resizebox{\linewidth}{!}{
    \begin{tabular}{c|ccc|ccc}
        \bottomrule 
        \multirow{2}{*}{Methods} & \multicolumn{3}{c|}{CAPE} & \multicolumn{3}{c}{4D-Dress} \\
        \cline{2-7}                          
        & CD $\downarrow$           & V2V $\downarrow$            & MPJPE $\downarrow$ & CD $\downarrow$ & V2V $\downarrow$ & MPJPE $\downarrow$ \\
        \Xhline{0.8pt}
        ETCH-X  (w/o mask)                             &    1.174                         &   1.975                          &       2.365             &       1.515          &   4.256               &     4.200        \\
        ETCH-X  (w/ mask)                             &   \textbf{1.160}                        &      \textbf{1.894}                       &    \textbf{2.266}                & \textbf{1.493}                 &     \textbf{4.169}             &    \textbf{4.083}                \\
        \toprule
    \end{tabular}
    }
    
    \label{tab:ablation_mask}
    \end{minipage}%
    
\vspace{-1.0 em}
\end{table}

\begin{figure}[t]
    \centering
    \includegraphics[width=0.95\linewidth]{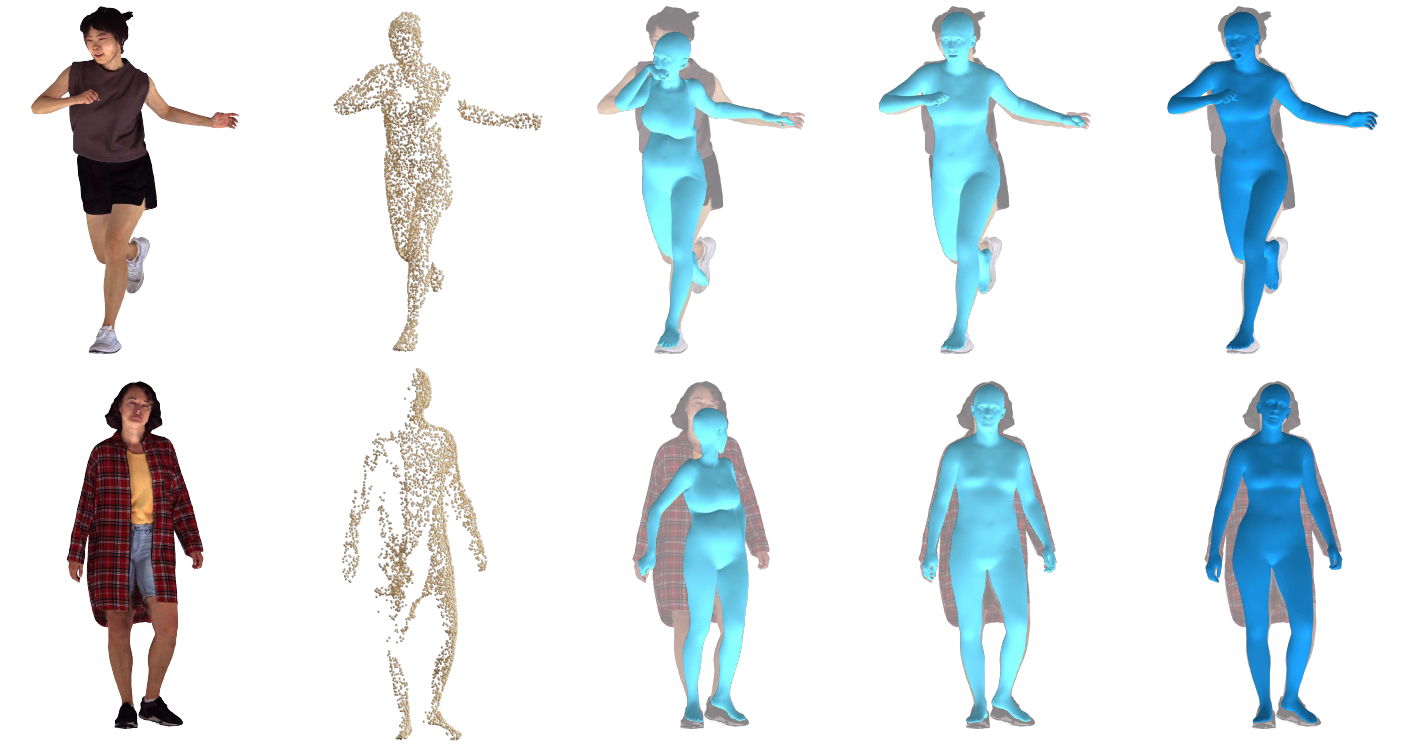}

    \begin{tabularx}
        {\linewidth}{ 
        >{\centering\arraybackslash}p{0.235\linewidth} >{\centering\arraybackslash}p{0.15\linewidth}
        >{\centering\arraybackslash}p{0.21\linewidth}>{\centering\arraybackslash}p{0.18\linewidth}>{\centering\arraybackslash}p{0.16\linewidth} }

        \scriptsize  Scan & \scriptsize  Partial Input& \scriptsize  ETCH-X w/o
        Aug & \scriptsize ETCH-X w/ Aug& \scriptsize  GT \\
    \end{tabularx}
    \vspace{-1.5 em}
    \caption{\scriptsize \textbf{Partial Augmentation 
    .} ETCH-X predicts better body poses with partial augmentation. 
    }
    \label{fig:partial_results}
\end{figure}

\Cref{subsec:exp_datasets} (Partial Data) details our single-view partial point cloud
simulation. During training, we randomly replace 50\% of full scans with partial
ones. As shown in~\cref{tab:ablation_partial}, partial augmentation improves
fitting performance by up to \textcolor{GreenColor}{72.5\%} on \ddress (V2V
metric) for partial inputs, while only slightly reducing accuracy on full scans
(maximum \textcolor{RedColor}{12.5\%} drop in V2V on \ddress). This highlights
the robustness of \modelname's modular design to partial inputs. Qualitative results
are shown in~\cref{fig:partial_results}.

\pheading{Disentangled Design.} The disentangled design of \modelname enables it
to effectively integrate simulated garment data and body pose libraries within a unified
framework. As demonstrated in \cref{tab:gen_data_comparison}, \modelname consistently
outperforms NICP and ETCH, 
trained exclusively on either AMASS or
CLOTH3D. By leveraging tightness vectors, \modelname achieves more accurate
undressing, particularly for loose garments in 4D-Dress where NICP often struggles.
In contrast, ETCH is limited in pose generalization due to the constrained pose diversity
in \clothd.

\begin{wrapfigure}
    {r}{0.5\textwidth}
    \vspace{-20pt}
    \centering
    \includegraphics[width=\linewidth]{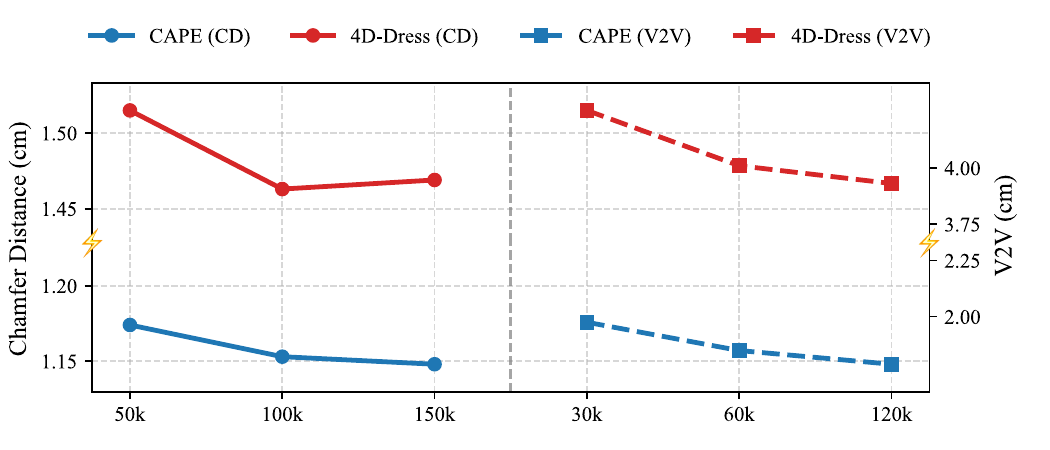}
    \caption{\scriptsize \textbf{Scaling Analysis of \modelname.} Increasing the
    amount of training data from \clothd~\cite{bertiche2020cloth3d} (left) and \amass~\cite{mahmood2019amass}
    (right) does not necessarily improve performance: tightness accuracy
    saturates, while pose robustness continues to increase.}
    \label{fig:gen_data_scaling}
    \vspace{-20pt}
\end{wrapfigure}

\pheading{Scaling Analysis.}
As discussed in \cref{sec:intro}, \modelname leverages both simulated garment data
(\clothd) and body pose libraries (\amass), enabling scalability across diverse
sources. \Cref{fig:gen_data_scaling} illustrates performance trends on \cape and
\ddress as the amount of training data increases. For tightness vectors derived
from \clothd, performance saturates rapidly, whereas adding more AMASS data
leads to steady improvements. We attribute this to the fact that predicting tightness
vectors depends on paired 3D scans, and the domain gap between real and
simulated data may constrain further gains. In contrast, expanding body pose
libraries allows the model to better cover test pose distributions, supporting continued
improvement.

\pheading{Tightness Masking.} As described in \cref{subsec:undress}, we introduce
a tightness mask to achieve more precise undressing by enforcing zero tightness on
exposed skin regions. 
Since CAPE dataset does not provide body segmentation, we perform tightness masking on CLOTH3D dataset.
Quantitative results in \cref{tab:ablation_mask} show that
lower Chamfer Distance (CD) errors indicate inner points that are more body-like,
which in turn leads to reduced V2V and MPJPE after fitting.

\begin{table}[t]
    \centering
    \caption{\textbf{Ablation Results of Hand Refinements.} 
    }
    \vspace{-1em}
    \renewcommand{\arraystretch}{1.2}
    \resizebox{\linewidth}{!}{
    \begin{tabular}{c|c|c|c|cc|cc|cc|cc}
        \bottomrule
        \multirow{3}{*}{Settings}  &\multirow{3}{*}{\makecell{Hand\\ LVD}} &\multirow{3}{*}{\makecell{Hand \\ Classifier}} &\multirow{3}{*}{\makecell{Hand Data \\Augmentation}} & \multicolumn{4}{c|}{CAPE}             & \multicolumn{4}{c}{4D-Dress}             \\
        \cline{5-12}     &&&         & \multicolumn{2}{c|}{V2V $\downarrow$} & \multicolumn{2}{c|}{MPJPE $\downarrow$} & \multicolumn{2}{c|}{V2V $\downarrow$} & \multicolumn{2}{c}{MPJPE $\downarrow$} \\
        \cline{5-12}     &&&         & All                                   & Hands                                   & All                                   & Hands                                 & All                            & Hands                           & All                            & Hands                           \\
        \Xhline{0.8pt} A.     &\xmark&\xmark&\xmark    & 1.550                                 & 2.607                                   & 1.948                                 & 2.467                                & 1.991                          & 3.417                           & 2.492                          & 3.367                           \\
        B.        &\cmark&\xmark& \xmark & 1.514                              & 2.417                                  & 1.835                                 & 2.277                                 & 1.963                          & 3.321                           & 2.444                          & 3.293                \\
        C.     &\cmark&\cmark& \xmark   & 1.493                                & 2.278                                  & 1.794                                 & 2.203                                 & 1.928                          & 3.167                           & 2.376                          & 3.125 \\
        ETCH-X     &\cmark&\cmark&\cmark     & \textbf{1.484}                                 & \textbf{2.215}                                    & \textbf{1.764}                                  & \textbf{2.148}                                  & \textbf{1.897}                           & \textbf{3.101}                            & \textbf{2.317}                           & \textbf{3.065}  \\
        \toprule
    \end{tabular}
    }
    
    \label{tab:ablation_hand}
    \vspace{-1.5 em}
\end{table}

\begin{figure}[t]
    \centering
    \includegraphics[width=\linewidth]{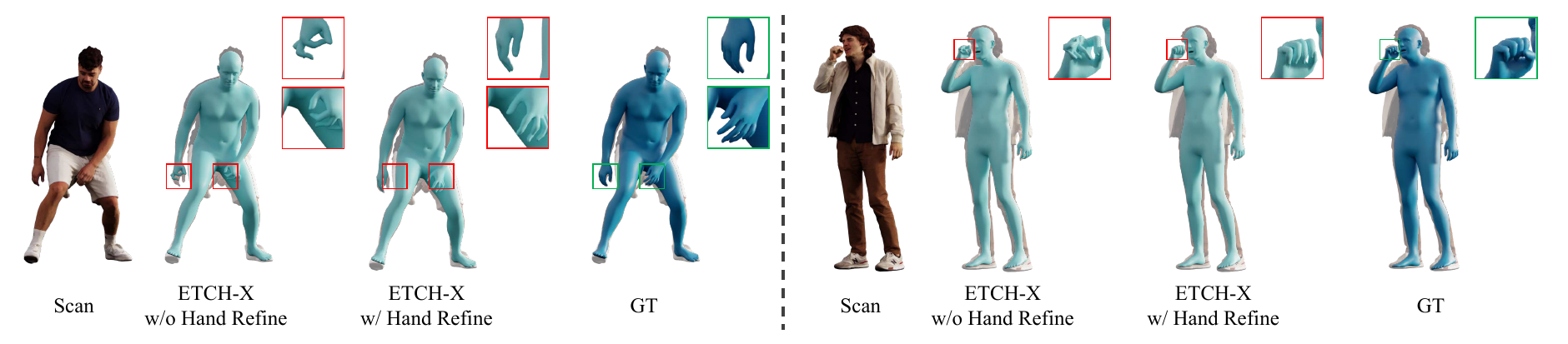}

    \vspace{-1.0 em}
    \caption{\scriptsize \textbf{Hand Refinement Results.} \modelname produce much better hand poses with hand refinement.}
    \label{fig:hand_comparison}
    \vspace{-2 em}
\end{figure}

\pheading{Hand Refinement.}
As described in \cref{subsec:dense_fit}, we adopt re-sampling to fit hand separately. The results under different settings are shown in \cref{tab:ablation_hand}, validating the effectiveness of our design. The visual comparison results in~\cref{fig:hand_refine} demonstrates the effect of hand refinement under conditions such as self-contact.

    \section{Conclusion}
\label{sec:conclusion}

We have presented \modelname, a novel two-stage pipeline for robustly fitting the \smplx body model to clothed 3D scans, regardless of garment type, clothing dynamics, body articulation, or partial observations. By decoupling the process into a masked undress stage and a dense fit stage, our framework remains flexible and scalable with composable synthetic data from diverse sources.
However, our approach has limitations, such as efficiency ($\sim$10 secs for a complete fitting pipeline), and simulated 3D garments have limited diversity.
Future work could focus on simulating more diverse 3D garments, and handling more complex scenarios like multi-person interactions and hybrid human-scene LiDAR capture, at real-time speed.

\section*{Acknowledgments}

We thank all the members of Endless AI Lab for their help and discussions. This work is funded by the Research Center for Industries of the Future (RCIF) at Westlake University, the Westlake Education Foundation.

    \clearpage
    \clearpage
    { \small \bibliographystyle{splncs04} \bibliography{main} }
\end{document}